\documentclass[10pt,twocolumn,letterpaper]{article}

\usepackage{iccv}
\usepackage{times}
\usepackage{epsfig}
\usepackage{graphicx}
\usepackage{amsmath}
\usepackage{amssymb}

\usepackage{multirow}
\usepackage{booktabs}
\usepackage{pifont}
\newcommand{\cmark}{\ding{51}}
\newcommand{\xmark}{\ding{55}}
\usepackage{color}
\usepackage[dvipsnames]{xcolor}
\usepackage{colortbl}

\usepackage[pagebackref=true,breaklinks=true,letterpaper=true,colorlinks,bookmarks=false]{hyperref}

\iccvfinalcopy 

\def\iccvPaperID{****} 

\ificcvfinal\fi

\begin{document}

\title{Incorporating Convolution Designs into Visual Transformers}

\author{Kun Yuan$^{\dag}$, \ Shaopeng Guo$^{\dag}$, \ Ziwei Liu$^{\ddag}$, \ Aojun Zhou$^{\dag}$, \ Fengwei Yu$^{\dag}$ and Wei Wu$^{\dag}$ \\
$^{\dag}$SenseTime Research, \ $^{\ddag}$S-Lab, Nanyang Technological University\\
{\tt\small \{yuankun,guoshaopeng,zhouaojun,yufengwei,wuwei\}@sensetime.com, ziwei.liu@ntu.edu.sg}
}

\maketitle
\ificcvfinal\thispagestyle{empty}\fi

\begin{abstract}

Motivated by the success of Transformers in natural language processing (NLP) tasks, there emerge some attempts (e.g., ViT and DeiT) to apply Transformers to the vision domain. However, pure Transformer architectures often require a large amount of training data or extra supervision to obtain comparable performance with convolutional neural networks (CNNs). To overcome these limitations, we analyze the potential drawbacks when directly borrowing Transformer architectures from NLP. Then we propose a new \textbf{Convolution-enhanced image Transformer (CeiT)} which combines the advantages of CNNs in extracting low-level features, strengthening locality, and the advantages of Transformers in establishing long-range dependencies. Three modifications are made to the original Transformer: \textbf{1)} instead of the straightforward tokenization from raw input images, we design an \textbf{Image-to-Tokens (I2T)} module that extracts patches from generated low-level features; \textbf{2)} the feed-froward network in each encoder block is replaced with a \textbf{Locally-enhanced Feed-Forward (LeFF)} layer that promotes the correlation among neighboring tokens in the spatial dimension; \textbf{3)} a \textbf{Layer-wise Class token Attention (LCA)} is attached at the top of the Transformer that utilizes the multi-level representations.

Experimental results on ImageNet and seven downstream tasks show the effectiveness and generalization ability of CeiT compared with previous Transformers and state-of-the-art CNNs, without requiring a large amount of training data and extra CNN teachers. Besides, CeiT models also demonstrate better convergence with $3\times$ fewer training iterations, which can reduce the training cost significantly\footnote{Code and models will be released upon acceptance.}.

\end{abstract}

\section{Introduction}

Transformers \cite{DBLP:conf/nips/VaswaniSPUJGKP17} have become the de-facto standard for natural language processing (NLP) tasks due to their abilities to model long-range dependencies and to train in parallel. Recently, there exist some attempts to apply Transformers to vision domains \cite{DBLP:conf/icml/ChenRC0JLS20,dosovitskiy2021an,DBLP:journals/corr/abs-2012-12877,DBLP:conf/eccv/CarionMSUKZ20,zhu2021deformable,DBLP:journals/corr/abs-2012-00364,DBLP:journals/corr/abs-2012-15840}, leading promising results in different tasks. Among them, Vision Transformer (ViT) \cite{dosovitskiy2021an} is the first pure Transformer architecture that is directly inherited from NLP, and applied to image classification. 
It obtains promising results compared to many state-of-the-art CNNs \cite{DBLP:conf/eccv/MahajanGRHPLBM18,DBLP:conf/cvpr/XieLHL20,DBLP:conf/eccv/KolesnikovBZPYG20}. 
But it relies heavily on the large amount of dataset of JFT-300M \cite{DBLP:conf/iccv/SunSSG17}, which limits the application in the scenarios with limited computing resources or labeled training data.
To alleviate the dependence on a large amount of data, the Data-efficient image Transformers (DeiT) \cite{DBLP:journals/corr/abs-2012-12877} introduce a CNN model as a teacher and applies knowledge distillation \cite{DBLP:journals/corr/HintonVD15} to improve the student model of ViT. Thus DeiT that is only trained on ImageNet can obtain satisfactory results. But the requirement of trained high-performance CNN models is a potential computation burden. Besides, the choice of teacher models, distillation types may affect the final performance. Therefore, we 
intend to design a new visual Transformer that can overcome these limitations.

\begin{figure}[t!]
  \centering
  \includegraphics[width=\linewidth]{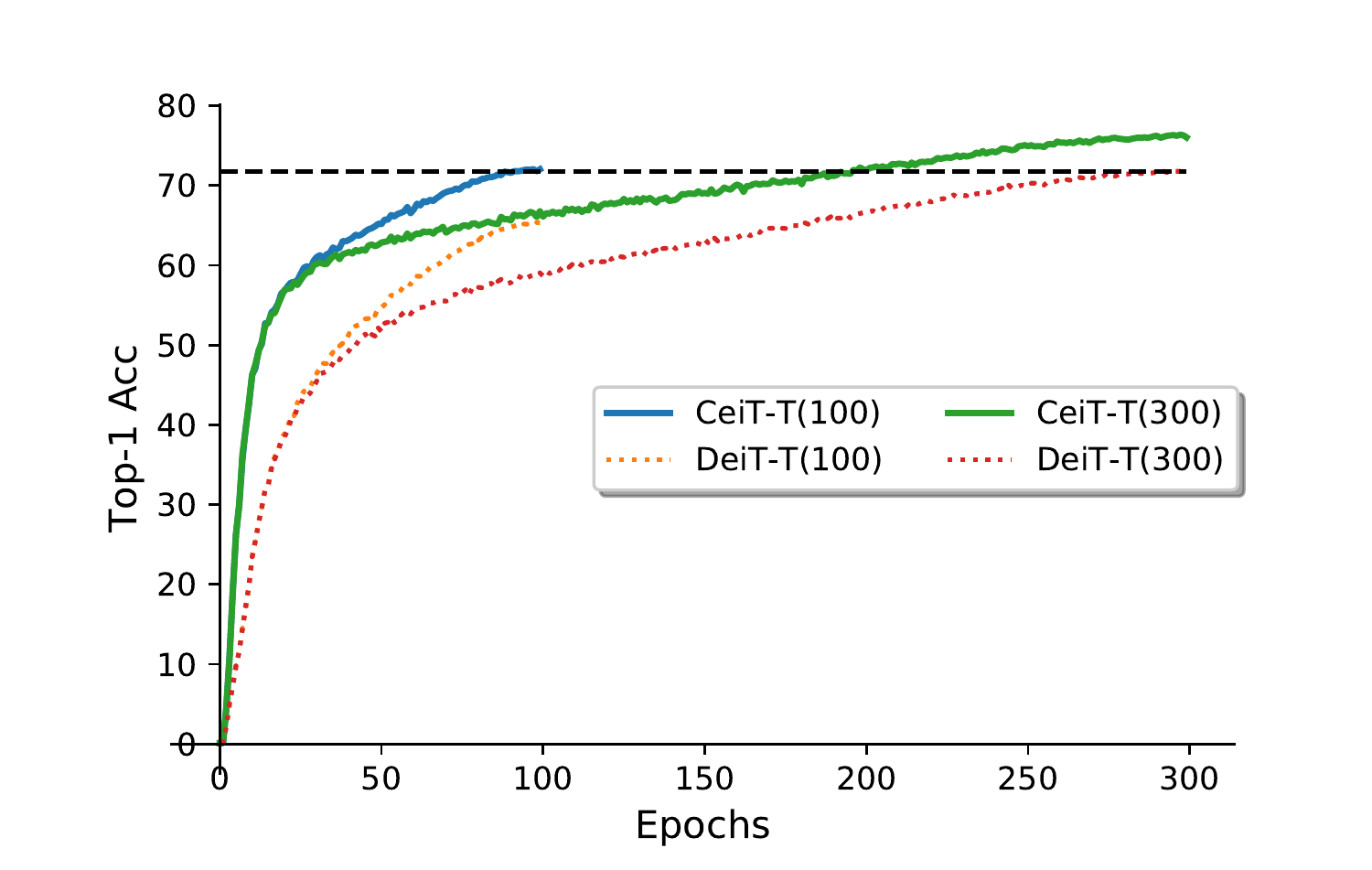}
  \caption{The fast convergence ability of CeiT models. CeiT models trained with 100 epochs obtain comparable results with DeiT models trained with 300 epochs. Other model settings are given in Table \ref{tab:convergence}.}
  \label{fig:curve}
\end{figure}

Some existing observations in these work can help us design desired architectures. In ViT, Transformer-based models underperform CNNs in the realm of $\sim$10M training samples. It claims that \textit{``Transformers lack some of the inductive biases inherent to CNNs, and therefore do not generalize well when trained on insufficient data"}. In DeiT, a CNN teacher gives better performance than using a Transformer one, which probably due to \textit{``the inductive bias inherited by the Transformer through distillation"}. These observations make us rethink whether it is appropriate to remove all convolutions from the Transformer. \textit{And should the inductive biases inherited in the convolution be forgotten?}

Looking back to the convolution, the main characteristics are translation invariance and locality \cite{DBLP:conf/nips/KrizhevskySH12,DBLP:journals/corr/SimonyanZ14a}. Translation invariance is relevant to the weight sharing mechanism, which can capture information about the geometry and topology in vision tasks \cite{DBLP:journals/neco/LeCunBDHHHJ89}. For the locality, it is a common assumption in visual tasks \cite{DBLP:conf/eccv/HeST10,DBLP:journals/spl/MihcakKRM99,DBLP:journals/tip/CriminisiPT04} that neighboring pixels always tend to be correlated. However, pure Transformer architectures do not fully utilize these prior biases that existed in images. \textit{First}, ViT performs direct tokenization of patches from the raw input image with a size of $16\times16$ or $32\times32$. It is difficult to extract the low-level features which form some fundamental structures in images (e.g. corners and edges). \textit{Second}, the self-attention modules concentrate on building long-range dependencies among tokens, ignoring the locality in the spatial dimension.

To address these problems, we design a {\textit{Convolution-enhanced image Transformer (CeiT)}} to combine the advantages of CNNs in extracting low-level features, strengthening locality, and the advantages of Transformers in associating long-range dependencies. Three modifications are made compared to the vanilla ViT. \textit{To solve the first problem}, instead of the straightforward tokenization from raw input images, we design an \textit{{Image-to-Tokens (I2T)}} module that extracts patches from generated low-level features, where patches are in a smaller size and then flattened into a sequence of tokens. Due to a well-designed structure, the I2T module does not introduce more computation costs. \textit{To solve the second problem}, the feed-froward network in each encoder block is replaced with a \textit{{Locally-enhanced Feed-Forward (LeFF)}} layer that promotes the correlation among neighboring tokens in the spatial dimension. \textit{To exploit the ability of self-attention}, a \textit{{Layer-wise Class token Attention (LCA)}} is attached at the top of the Transformer that utilizes the multi-level representations to improve the final representation.
In summary, our \textbf{contributions} are as follows:
\begin{itemize}
    \item {We design a new visual Transformer architecture namely \textit{Convolution-enhanced image Transformer (CeiT)}. It combines the advantages of CNNs in extracting low-level features,  strengthening locality, and the advantages of Transformers in establishing long-range dependencies.}
    \item{Experimental results on ImageNet and seven downstream tasks show the effectiveness and generalization ability of CeiT compared with previous Transformers and state-of-the-art CNNs, without requiring a large amount of training data and extra CNN teachers. For example, with a similar model size as ResNet-50, CeiT-S obtains a Top-1 accuracy of $82.0\%$ on ImageNet. And the result boosts into $83.3\%$ when fine-tuned in the resolution of $384\times 384$.}
    \item As shown in Figure \ref{fig:curve}, CeiT models demonstrate better convergence than pure Transformer models with $3\times$ fewer training iterations, which can reduce the training cost significantly.
\end{itemize}


\section{Related Work}

\paragraph{Transformer in Vision.} 
iGPT \cite{DBLP:conf/icml/ChenRC0JLS20} first introduce transformer to auto-regressively predict pixels, and obtaining pre-trained models without incorporating knowledge of the content in 2D images. However, it can only achieve reasonable performance in a tiny dataset (CIFAR10) with an extremely large model (1.4B). 
Recently, ViT \cite{dosovitskiy2021an} successfully makes standard Transformer scalable for image classification. It reshapes the images into a series of $16\times 16$ patches as input tokens. However, ViT can only get comparable performance with state-of-the-art CNNs when trained on very large datasets.
DeiT \cite{DBLP:journals/corr/abs-2012-12877} augment ViT by introducing a mimic token and adopt knowledge distillation to mimic the output of a CNN teacher, which can obtain satisfactory results without training on large scale dataset.
Some work also exploit efficient Transformers that can be trained in ImageNet directly, including LambdaNetworks \cite{bello2021lambdanetworks}, 
T2T-ViT \cite{DBLP:journals/corr/abs-2101-11986} and PVT \cite{DBLP:journals/corr/abs-2102-12122}.
Besides, recent work also apply Transformers to various vision tasks, including object detection \cite{DBLP:conf/eccv/CarionMSUKZ20,zhu2021deformable}, segmentation \cite{DBLP:journals/corr/abs-2011-14503}, image enhancement \cite{DBLP:journals/corr/abs-2012-00364,DBLP:conf/cvpr/YangYFLG20} and video processing \cite{DBLP:conf/eccv/ZengFC20,DBLP:conf/cvpr/ZhouZCSX18}.


\paragraph{Hybrid Models of Convolution and Self-attention.}

To utilize the advantages of self-attention in building long-range dependencies, some work introduces attention modules into CNNs~\cite{DBLP:conf/eccv/ZhaoZLSLLJ18,DBLP:conf/cvpr/YangHGDS16,DBLP:conf/cvpr/0004GGH18,DBLP:conf/iccv/BelloZLVS19,DBLP:conf/nips/ChenKLYF18, DBLP:conf/iccv/HuZXL19,DBLP:conf/eccv/WooPLK18}. Among these works, the Non-local network \cite{DBLP:conf/cvpr/0004GGH18} insert non-local layers into the last several blocks of ResNet~\cite{DBLP:conf/cvpr/HeZRS16} and improve the performance on video recognition and instance segmentation. CCNet \cite{DBLP:conf/iccv/HuangWHHW019} attaches a criss-cross attention module at the top of a segmentation network. 
SASA \cite{DBLP:conf/nips/ParmarRVBLS19}, SANet \cite{DBLP:conf/cvpr/ZhaoJK20} and Axial-SASA \cite{DBLP:conf/eccv/WangZGAYC20} propose to replace all convolutional layers by self-attention module to form a stand-alone self-attention network.
Recent work also combines Transformers with CNNs. DETR \cite{DBLP:conf/eccv/CarionMSUKZ20} uses Transformer blocks outside the CNN backbone with the motivation to get rid of region proposals and non-maximal suppression for simplicity. 
ViLBERT \cite{DBLP:conf/nips/LuBPL19} and VideoBERT \cite{DBLP:conf/iccv/SunMV0S19} construct cross-modality models using CNN and BERT. 
Different from the above methods, CeiT incorporates convolutional designs into the basic building blocks of Transformer to inherit the inductive bias in CNNs, which a more elaborate design.

\section{Methodology}\label{sec:method}

Our CeiT is designed based on the ViT. First, we give a brief overview of the basic components of ViT in section \ref{sec:vit_overview}. Next, we introduce three modifications that incorporate convolution designs and benefit visual Transformers, including an \textit{Image-to-Tokens (I2T)} module in section \ref{sec:i2t}, a \textit{Locally-enhanced FeedForwad (LeFF)} module in section \ref{sec:leff} and 
a \textit{Layer-wise Class token Attention (LCA)} module in section \ref{sec:lca}. Last, we analyze the computation complexity of these proposed modules in section \ref{sec:complexity}.

\begin{figure}[t!]
  \centering    
  \includegraphics[width=\linewidth]{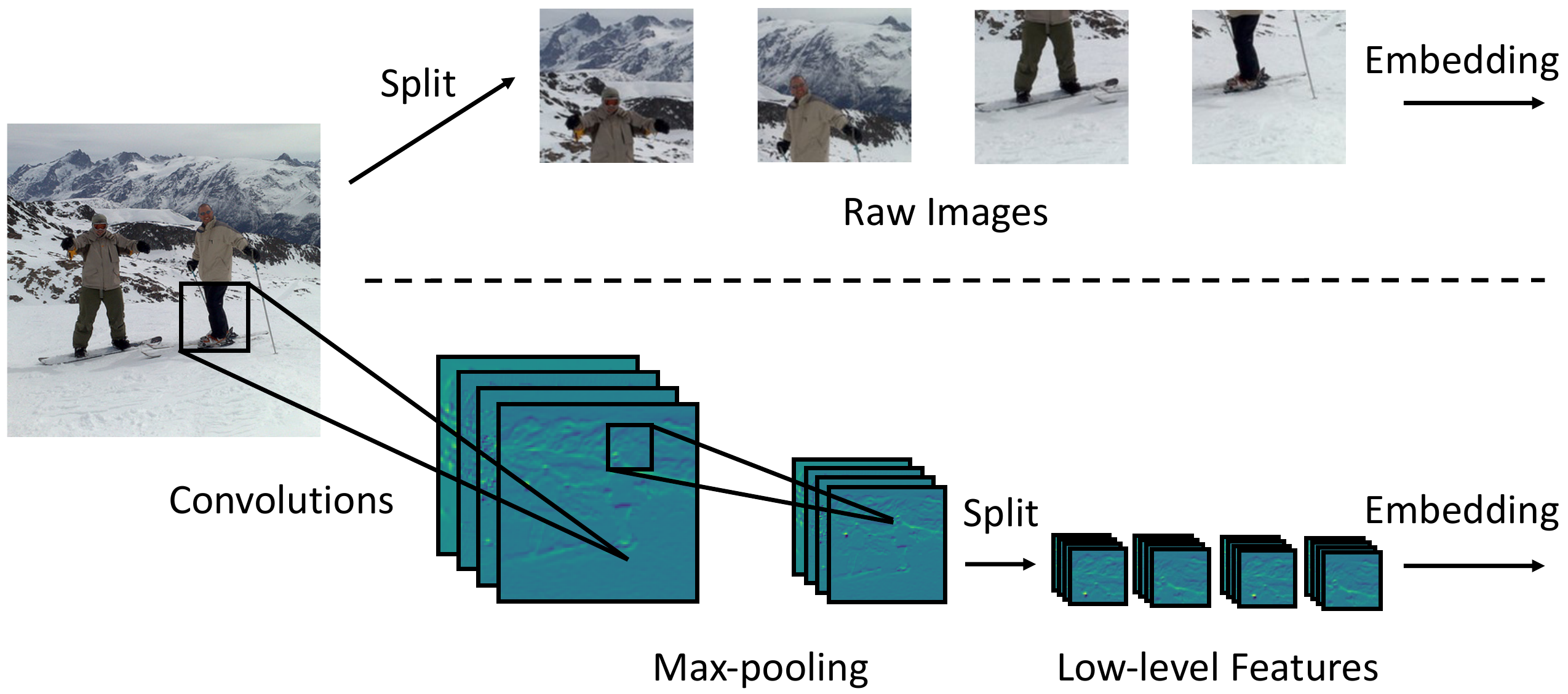}
  \caption{Comparisons of different tokenization methods. The upper one extracts patches from raw input images. The below one (I2T) uses the low-level features generated by a convolutional stem.}
  \label{fig:i2t}
\end{figure}
\subsection{Revisiting Vision Transformer}\label{sec:vit_overview}

We first revisit the basic components in ViT, including tokenization, encoder blocks, multi-head self-attention (MSA) layers, and feed-forward network (FFN) layers.

\paragraph{Tokenization.} The standard Transformer \cite{DBLP:conf/nips/VaswaniSPUJGKP17} receives a sequence of token embeddings as input. To handle 2D images, ViT reshapes the image $\mathbf{x}\in \mathbb{R}^{H\times W\times 3}$ into a sequence of flattened 2D patches $\mathbf{x}_p \in \mathbb{R}^{N \times (P^2 \cdot 3)}$, where $(H, W)$ is the resolution of the original image, $3$ is the number of channels of RGB images, $(P, P)$ is the resolution of each image patch, and $N=HW/P^2$ is the resulting number of patches, which also serves as the effective input sequence length for the Transformer. And these patches are flattened and mapped to latent embeddings with a size of $C$. Then an extra class token is added to the sequence and serves as the image representation, resulting in the input of sequence with a size of $\mathbf{x}_t \in \mathbb{R}^{(N+1)\times C}$.

In practice, ViT splits each image with a patch size of $16 \times 16$ or $32 \times 32$. But the straightforward tokenization of input images with large patches may have two limitations: 1) it is difficult to capture low-level information in images (such as edges and corners); 2) large kernels are over-parameterized and are often hard to optimize, thus requires much more training samples or training iterations.

\paragraph{Encoder blocks.} ViT is composed of a series of stacked encoders. Each encoder has two sub-layers of MSA and FFN. A residual connection \cite{DBLP:conf/cvpr/HeZRS16} is employed around each sub-layer, followed by layer normalization (LN)~\cite{DBLP:journals/corr/BaKH16}. The output for each encoder is:
\begin{equation}
    \mathbf{y} = \mbox{LN}(\mathbf{x}^{\prime} + \mbox{FFN}(\mathbf{x}^{\prime})), \mbox{and} \ \mathbf{x}^{\prime} = \mbox{LN}(\mathbf{x} + \mbox{MSA}(\mathbf{x}))
\end{equation}

Different from CNNs where feature maps are down-sampled at the beginning of each stage, the length of tokens is not reduced in different Encoder blocks. The effective receptive field cannot be expanded efficiently, which may affect the efficiency of optimization in visual Transformers.

\paragraph{MSA.} For a self-attention (SA) module, the sequence of input tokens $\mathbf{x}_t \in \mathbb{R}^{(N+1)\times C}$ are linear transformed into \texttt{qkv} spaces, $i.e.,$ queries $\mathbf{Q}\in \mathbb{R}^{(N+1)\times C}$, keys $\mathbf{K}\in \mathbb{R}^{(N+1)\times C}$ and values $\mathbf{V}\in \mathbb{R}^{(N+1)\times C}$. Then a weighted sum over all values in the sequence is computed through:
\begin{equation}
    \mbox{Attention}(\mathbf{Q},\mathbf{K},\mathbf{V}) = \mbox{softmax}(\frac{\mathbf{QK}^T}{\sqrt{C}})\mathbf{V}
\end{equation}
And a linear transformation is performed to the weighted values. MSA is an extension of SA. It splits queries, keys, and values for $\mathbf{h}$ times and performs the attention function in parallel, then projects their concatenated outputs. 

Through computing dot-product, the similarity between different tokens is calculated, resulting in long-range and global attention. And a linear aggregation is performed for corresponding values $\mathbf{V}$.

\begin{figure*}[t!]
  \centering
  \includegraphics[width=0.85\linewidth]{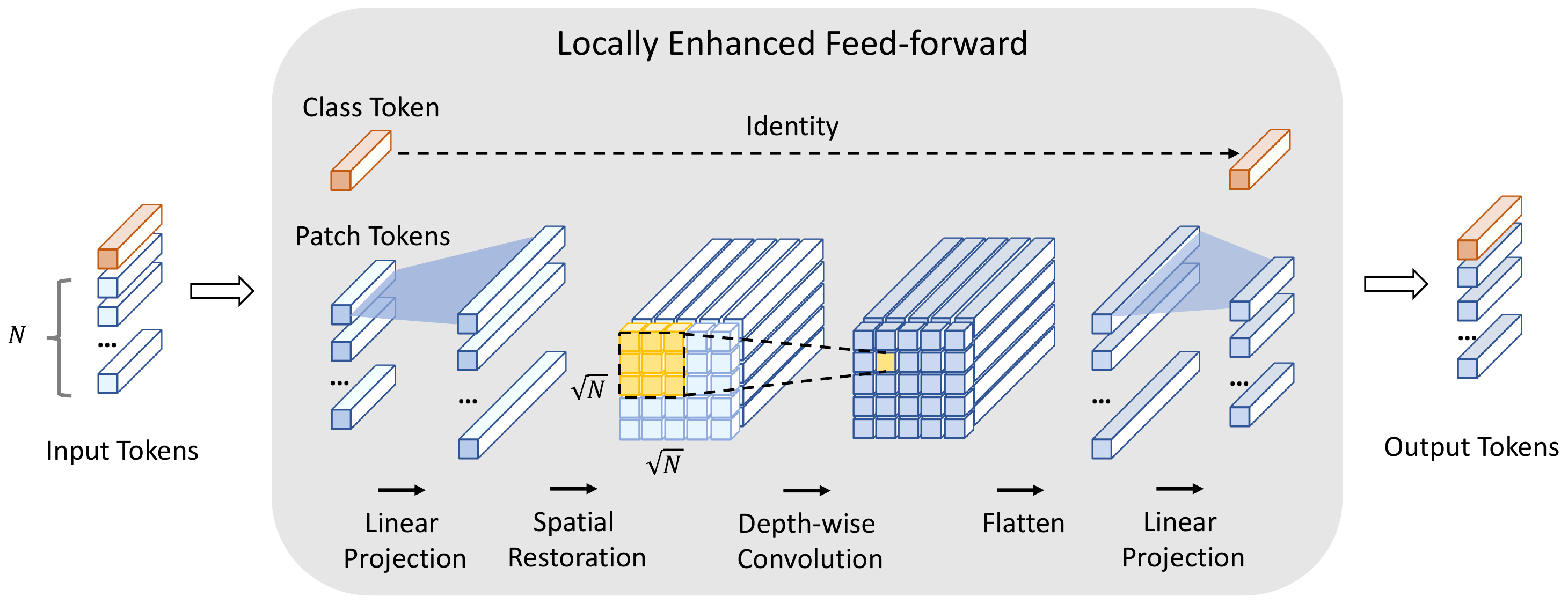}
  \caption{Illustration of the Locally-enhanced Feed-Forward module. First, patch tokens are projected into a higher dimension. Second, they are restored to ``images" in the spatial dimension based on the original positions. Third, a depth-wise convolution is performed on the restored tokens as shown in the yellow region. Then the patch tokens are flattened and projected to the initial dimension. Besides, the class token conducts an identical mapping.}
  \label{fig:leff}
\end{figure*}

\paragraph{FFN.} FFN performs point-wise operations, which are applied to each token separately. It consists of two linear transformations with a non-linear activation in between:
\begin{equation}
    \mbox{FFN}(\mathbf{x}) = \sigma(\mathbf{x}\mathbf{W}_1+\mathbf{b}_1)\mathbf{W}_2+\mathbf{b}_2
\end{equation}
where $\mathbf{W}_1 \in \mathbb{R}^{C\times K}$ is the weight of the first layer, projecting each token into a higher dimension $K$. And $\mathbf{W}_2 \in \mathbb{R}^{K\times C}$ is the weight of the second layer. $\mathbf{b}_1 \in \mathbb{R}^{K}$ and $\mathbf{b}_2 \in \mathbb{R}^{C}$ are the biases. And $\sigma(\cdot)$ is the non-linear activation of GELU \cite{DBLP:journals/corr/HendrycksG16} in ViT.

Complementary to the MSA module, the FFN module performs dimensional expansion/reduction and non-linear transformation on each token, thereby enhancing the representation ability of tokens. However, the spatial relationship among tokens, which is important in vision, is not considered. This leads that the original ViT needs a mass of training data to learn these inductive biases.

\subsection{Image-to-Tokens with Low-level Features}\label{sec:i2t}

To solve the above-mentioned problems in tokenization, we propose a simple but effective module named as \textit{Image-to-Tokens (I2T)} that extracts patches from feature maps instead of raw input images. As shown in Figure \ref{fig:i2t}, the I2T module is an lightweight stem that consists of a convolutional layer and a max-pooling layer. Ablation studies also suggest that a BatchNorm layer following the convolution layer benefits the training process. It can be denoted as:
\begin{equation}
    \mathbf{x}^{\prime} = \mbox{I2T}(\mathbf{x}) = \mbox{MaxPool}(\mbox{BN}(\mbox{Conv}(\mathbf{x})))
\end{equation}
where $\mathbf{x}^{\prime} \in \mathbb{R}^{\frac{H}{S}\times \frac{W}{S} \times D}$, $S$ is the stride w.r.t the raw input images, and $D$ is the number of enriched channels. Then the learned feature maps are extracted into a sequence of patches in the spatial dimension. To keep the number of generated tokens consistent with ViT, we shrink the resolution of patches into $(\frac{P}{S}, \frac{P}{S})$. In practice, we set $S=4$.

I2T fully utilizes the advantage of CNNs in extracting low-level features and reduces the training difficulty of embedding by shrinking the patch size. This is also different from the hybrid type of Transformer proposed in ViT, where a regular ResNet-50 is used to extract high-level features from the last two stages. Our I2T is much lighter.



\subsection{Locally-Enhanced Feed-Forward Network}\label{sec:leff}

To combine the advantage of CNNs to extract local information with the ability of Transformer to establish long-range dependencies, we propose a \textit{Locally-enhanced Feed-Forward Network (LeFF)} layer. In each Encoder block, we keep the MSA module unchanged, remaining the ability to capture global similarities among tokens. Instead, the original feed-forward network layer is replaced with the LeFF. The structure is given in Figure \ref{fig:leff}.

A LeFF module performs following procedures. \textit{First}, given tokens $\mathbf{x}_t^{h}\in \mathbb{R}^{(N+1)\times C}$ generated from the preceding MSA module, we split them into patch tokens $\mathbf{x}_p^{h}\in \mathbb{R}^{(N+1)\times C}$ and a class token $\mathbf{x}_c^{h}\in \mathbb{R}^{C}$ accordingly. A linear projection is conducted to expand the embeddings of patch tokens to a higher dimension of $\mathbf{x}_p^{l_1}\in \mathbb{R}^{N\times (e\times C)}$, where $e$ is the expand ratio. \textit{Second}, the patch tokens are restored to ``images" of $\mathbf{x}_p^{s}\in \mathbb{R}^{\sqrt{N}\times \sqrt{N}\times (e\times C)}$ on spatial dimension based on the position relative to the original image. \textit{Third}, we perform a depth-wise convolution with kernel size of $k$ on these restored patch tokens, enhancing the representation correlation with neighboring $k^2-1$ tokens, obtaining $\mathbf{x}_p^{d}\in \mathbb{R}^{\sqrt{N}\times \sqrt{N}\times (e\times C)}$. \textit{Fourth}, these patch tokens are flattened into sequence of $\mathbf{x}_p^{f}\in \mathbb{R}^{N\times (e\times C)}$. \textit{Last}, the patch tokens are projected to the initial dimension with $\mathbf{x}_p^{l_2}\in \mathbb{R}^{N\times C}$, and concatenated with the class token, resulting in $\mathbf{x}_t^{h+1}\in \mathbb{R}^{(N+1)\times C}$. Following each linear projection and depth-wise convolution, a BatchNorm and a GELU is added. These procedures can be noted as:
\begin{align}
	\mathbf{x}_c^h, \mathbf{x}_p^h &= \mbox{Split}(\mathbf{x}_t^h) \\
	\mathbf{x}_p^{l_1} &= \mbox{GELU}(\mbox{BN}(\mbox{Linear1}(\mathbf{x}_p^h))) \\
	\mathbf{x}_p^{s} &= \mbox{SpatialRestore}(\mathbf{x}_p^{l_1}) \\
	\mathbf{x}_p^{d} &= \mbox{GELU}(\mbox{BN}(\mbox{DWConv}(\mathbf{x}_p^{s}))) \\
	\mathbf{x}_p^{f} &= \mbox{Flatten}(\mathbf{x}_p^{d}) \\
	\mathbf{x}_p^{l_2} &= \mbox{GELU}(\mbox{BN}(\mbox{Linear2}(\mathbf{x}_p^f))) \\
	\mathbf{x}_t^{h+1} &= \mbox{Concat}(\mathbf{x}_c^h, \mathbf{x}_p^{l_2})
\end{align}

\subsection{Layer-wise Class-Token Attention}\label{sec:lca}

In CNNs, as the network deepens, the receptive field of the feature map increases. Similar observations are also found in ViT, whose ``attention distance" increases with depth. Therefore, feature representations will be different at different layers. To integrate information across different layers, we design a \textit{Layer-wise Class-token Attention (LCA)} module. Unlike the standard ViT that takes the class token $\mathbf{x}_c^{(L)}$ at the last $L$-th layer as the final representation, LCA makes attention over class tokens at different layers.

\begin{figure}[t!]
  \centering
  \includegraphics[width=0.85\linewidth]{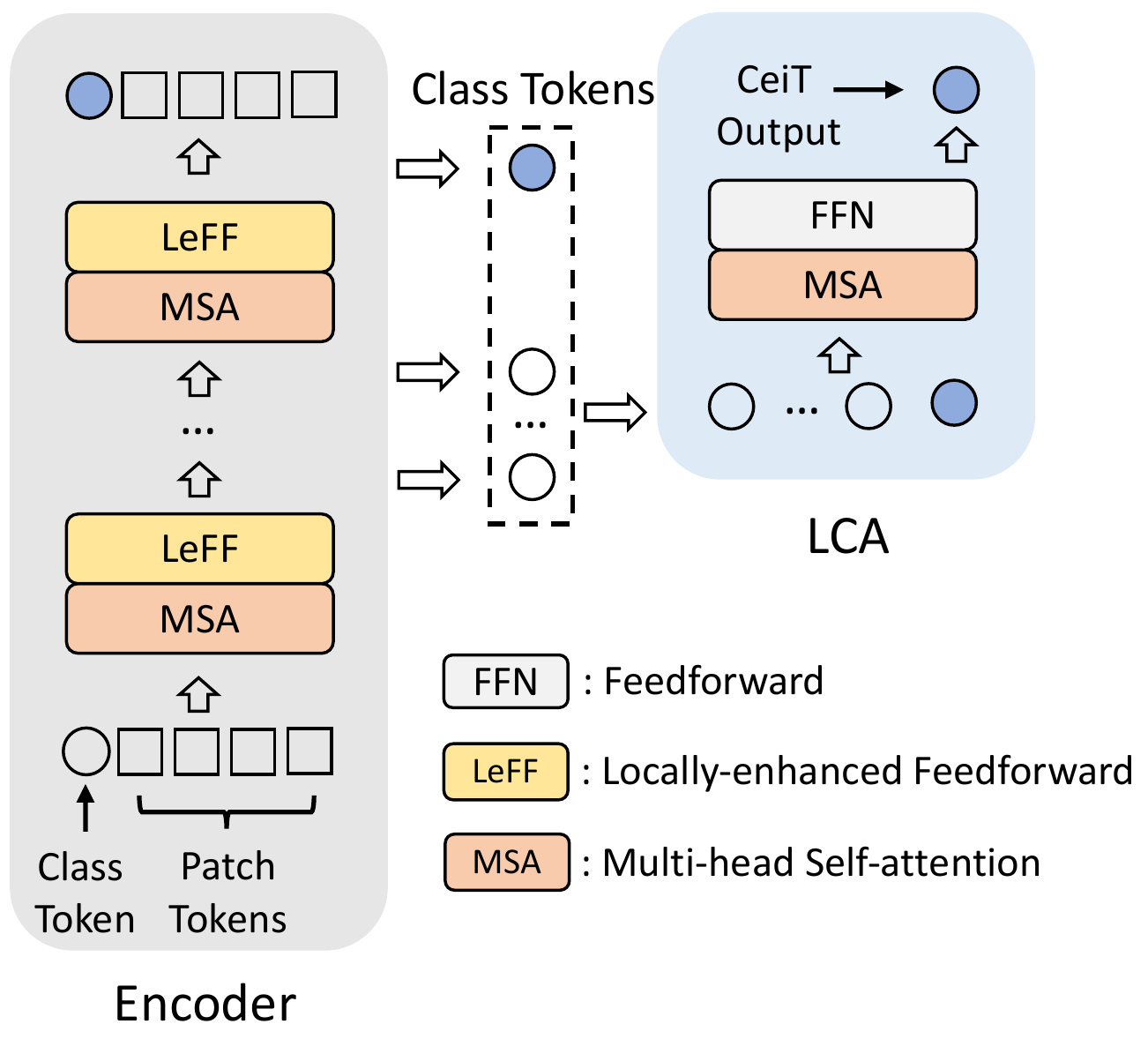}
  \caption{The proposed Layer-wise Class-token Attention block. It integrates information across different layers through receiving a sequence of class tokens as inputs.}
  \label{fig:lca}
\end{figure}

As shown in Figure \ref{fig:lca}, LCA gets a sequence of class tokens as the input, which can be denoted as $\mathbf{X}_c = [\mathbf{x}_c^{(1)}, \cdots, \mathbf{x}_c^{(l)}, \cdots, \mathbf{x}_c^{(L)}]$, where $l$ denotes the layer depth. LCA follows the standard Transformer block, which contains a MSA and a FFN layer. Particularly, it only computes unidirectional similarities between the $L$-th class token $\mathbf{x}_c^{(L)}$ and other class tokens. These modifications reduce the complexity of computing attention from $O(n^2)$ to $O(n)$. And the corresponding value of $\mathbf{x}_c^{(L)}$ is aggregated with others through attention. Then the aggregated value is sent into a FFN layer, resulting in the final representations of $\mathbf{x}_c^{(L)^{\prime}}$.

\subsection{Computational Complexity Analysis}\label{sec:complexity}

We analyze the extra computational complexity (in terms of FLOPs) brought by our modifications. Generally, with a small increase in computational cost, our CeiT model can efficiently combine the advantage of CNNs and Transformers, resulting in higher performance and better convergence.

\paragraph{I2T vs Original.} The type of tokenization affects the computational cost of embedding. For the original one with a patch size of $16 \times 16$, the FLOPs are $3C(HW)^2$. For I2T, the FLOPs are consist of two parts, including feature generation and embedding. In this paper, the generated features are $4\times$ smaller than the input. And the detailed architecture of I2T is given in section \ref{sec:exp_setting}. The total FLOPs of I2T are $(\frac{147}{4}+\frac{9}{64})DHW + \frac{1}{64}DCHW$. For a ViT-B/16 model, the ratio between I2T and the original one is around $1.1$. In this way, the extra computational cost is negligible.

\paragraph{LeFF vs FFN.} In a FFN layer with $e=4$, the FLOPs are $8(N+1)C^2$. The main extra computation cost of LeFF is introduced by the depth-wise convolution, whose FLOPs are $4k^2N^2C$. The increase of FLOPs is small since $O((N+1)C^2)\gg O(N^2C)$ in practice as given in Table \ref{tab:arch}.

\paragraph{LCA vs Encoder Block.} Compared with the standard Encoder block, the LCA only computes attention over the $L$-th class token. Both the computation cost in MSA and FFN has been reduced to $\frac{1}{N}$. The cost can be ignored compared with the other 12 encoder blocks.



	

\section{Experiments}

We perform extensive experiments to demonstrate the effectiveness of our proposed CeiT. In section \ref{sec:exp_setting}, we give the details of used visual datasets and training settings. In section \ref{sec:exp_imagenet}, we compare CeiT with other state-of-the-art architectures including CNNs and Transformers in ImageNet.  In section \ref{sec:exp_transfer}, we transfer CeiT models trained on ImageNet to other benchmark datasets, showing the strong generalization ability. In section \ref{sec:exp_ablation}, we conduct ablation studies on our modifications. In section \ref{sec:exp_fast}, we show the fast convergence ability of our CeiT models.

\begin{table*}[t]
    \centering
    \small
    \caption{Variants of our CeiT architecture. The FLOPs are calculated for images at resolution $224\times 224$. And $7ks2$ means a convolution/pooling with kernel size of 7 and stride of 2.}
    \begin{tabular}{c|c|c|c|c|c|c|c|c|c|c}
    \toprule
    \multirow{2}{*}{Model} & \multicolumn{3}{c}{I2T} & encoder & embedding & \multirow{2}{*}{heads} & \multicolumn{2}{c}{LeFF} & Params & FLOPs \\
                           & conv & maxpool & channels & blocks & dimension & & $e$ & $k$ & (M) & (G) \\
    \midrule
    CeiT-T & $k7s2$ & $k3s2$ & 32 & 12 & 192 & 3 & 4 & 3 & 6.4 & 1.2 \\
    CeiT-S & $k7s2$ & $k3s2$ & 32 & 12 & 384 & 6 & 4 & 3 & 24.2 & 4.5 \\
    CeiT-B & $k7s2$ & $k3s2$ & 32 & 12 & 768 & 12 & 4 & 3 & 86.6 & 17.4 \\
    \bottomrule
    \end{tabular}
    \label{tab:arch}
\end{table*}

\begin{table*}[t]
   \centering
   \small
   \caption{The hyper-paramters that varies depends on tasks. The training and fine-tuning on ImageNet  in our experienet adopt the same setting in DeiT. We use the same settings for finetuning on different downstream datasets.}
   \begin{tabular}{c|c|c|c|c|c|c|c|c|c}
   \toprule
        \multirow{2}{*}{Task} & \multirow{2}{*}{dataset} & input & \multirow{2}{*}{Epochs} & batch & learning & LR & warmup & weight & repeated  \\
        & & size & & size & rate & scheduler & epoch & decay & aug \cite{DBLP:conf/cvpr/HofferBHGHS20} \\
        \midrule
        training & ImageNet & 224 & 300 & 1024 & 1e-3 & cosine & 5  & 0.05 & \cmark \\
        fine-tuning & ImageNet & 384 & 30 & 1024 & 5e-6 & constant & 0 & 1e-8 & \cmark \\
        transferring & downstream & 224$\&$384 & 100 & 512 & 5e-4 & cosine & 2 &  1e-8 & \xmark \\
   \bottomrule
   \end{tabular}
   \label{tab:hyper}
\end{table*}

\subsection{Experimental Settings}\label{sec:exp_setting}

\begin{table}[t]
    \centering
    \small
    \caption{Details of used visual datasets.}
    \begin{tabular}{c|ccc}
    \toprule
         dataset & classes & train data & val data  \\
         \midrule
         ImageNet & 1000 & 1,281,167 & 50000 \\
         \midrule
         iNaturalist2018 & 8142 & 437513 & 24426 \\
         iNaturelist2019 & 1010 & 265240 & 3003 \\
         \midrule
         Standford Cars & 196 & 8133 & 8041 \\
         Oxford-102 Followers & 102 & 2040 & 6149 \\
         Oxford-IIIT-Pets & 37 & 3680 & 3669 \\
         \midrule
         CIFAR100 & 100 & 50000 & 10000 \\
         CIFAR10 & 10 & 50000 & 10000 \\
         \bottomrule  
    \end{tabular}
    \label{tab:dataset}
\end{table}

\paragraph{Network Architectures.} We build our CeiT architectures by following the basic configurations of ViT and DeiT. The details are given in Table \ref{tab:arch}. The I2T module consists of a convolutional layer with a kernel size of 7 and a stride of 2, generating enriched channels of $32$. And a BatchNorm layer is added for stable training. Then a max-pooling layer with a kernel size of 3 and a stride of 2 is followed, resulting in feature maps with $4\times$ smaller than the input image. Compared to the patch size of $16\times 16$ in ViT, we use a patch size of $4\times 4$ in generating a sequence of tokens. We follow the standard setting in the number of the depth of 12. For the LeFF module, we set the expand ratio $e$ to be 4. And the kernel size for the depth-wise convolution is $3\times 3$. For the LCA module, the number of heads and the ratio of MLP follow those of the standard Encoder blocks.

\paragraph{Implementation Details.} All of our experiments are performed on the NVIDIA Tesla V100 GPUs. We adopt the same training strategy in DeiT. We list the detailed settings for training, fine-tuning, and transfer learning in Table \ref{tab:hyper}.

\paragraph{Datasets.} Instead of using a large-scale training dataset of JFT300M or ImageNet22K, we adopt the mid-sized ImageNet \cite{DBLP:journals/ijcv/RussakovskyDSKS15} dataset. It consists of 1.2M training images belonging to 1000 classes, and 50K validation images. Besides, we also test on some downstream benchmarks to evaluate the transfer ability of our trained CeiT models. These datasets consist different scenes, including fine-grained recognition (Standford Cars \cite{DBLP:conf/iccvw/Krause0DF13}, Oxford-102 Followers \cite{DBLP:conf/icvgip/NilsbackZ08} and Oxford-IIIT-Pets \cite{DBLP:conf/cvpr/ParkhiVZJ12}), long-tailed classification (iNaturalist18 \cite{DBLP:journals/corr/HornASSAPB17}, iNaturalist19 \cite{DBLP:journals/corr/HornASSAPB17}) and superordinate level classification (CIFAR10 \cite{Krizhevsky09learningmultiple}, CIFAR100 \cite{Krizhevsky09learningmultiple}). The details are given in Table \ref{tab:dataset}.

\subsection{Results on ImageNet}\label{sec:exp_imagenet}

\begin{table*}[t]
    \centering
    \footnotesize
    \caption{Accuracies on ImageNet and ImageNet Real of CeiT and of several SOTA CNNs and Transformers, for models trained with no extra data. The notation $\uparrow384$ means the model is fine-tuned on the resolution of $384\times384$. CeiT models outperform CNNs with similar FLOPs. Directly trained in ImageNet, CeiT models also achieve higher performances than DeiT models that are trained with extra CNN teachers.}
    \begin{tabular}{c|c|ccc|cc|c}
        \toprule
        \multirow{2}{*}{Group} &\multirow{2}{*}{Model} & FLOPs & Params & input
        & \multicolumn{2}{c|}{ImageNet} & Real \\
        & & (G) & (M) & size & Top-1 & Top-5 & Top-1 \\
        \midrule
        \multirow{12}{*}{CNNs}&ResNet-18~\cite{DBLP:conf/cvpr/HeZRS16} & 1.8 & 11.7 & 224 & 70.3 & 86.7 & 77.3 \\
        &ResNet-50~\cite{DBLP:conf/cvpr/HeZRS16} & 4.1 & 25.6 & 224 & 76.7 & 93.3 & 82.5 \\
        &ResNet-101~\cite{DBLP:conf/cvpr/HeZRS16} & 7.8 & 44.5 & 224 & 78.3 & 94.1 & 83.7 \\
        &ResNet-152~\cite{DBLP:conf/cvpr/HeZRS16} & 11.5 & 60.2 & 224 & 78.9 & 94.4 & 84.1 \\
        \cmidrule{2-8}
        &EfficientNet-B0~\cite{DBLP:conf/icml/TanL19} & 0.4 & 5.3 & 224 & 77.1 & 93.3 & 83.5 \\
        &EfficientNet-B1~\cite{DBLP:conf/icml/TanL19} & 0.7 & 7.8 & 240 & 79.1 & 94.4 & 84.9 \\
        &EfficientNet-B2~\cite{DBLP:conf/icml/TanL19} & 1.0 & 9.1 & 260 & 80.1 & 94.9 & 85.9 \\
        &EfficientNet-B3~\cite{DBLP:conf/icml/TanL19} & 1.8 & 12.2 & 300 & 81.6 & 95.7 & 86.8 \\
        &EfficientNet-B4~\cite{DBLP:conf/icml/TanL19} & 4.4 & 19.3 & 380 & 82.9 & 96.4 & 88.0 \\
        \cmidrule{2-8}
        &RegNetY-4GF~\cite{radosavovic2020designing} & 4.0 & 20.6 & 224 & 80.0 & - & 86.4 \\
        &RegNetY-8GF~\cite{radosavovic2020designing} & 8.0 & 39.2 & 224 & 81.7 & - & 87.4 \\
        \midrule
        \multirow{22}{*}{Transformers}& ViT-B/16~\cite{dosovitskiy2021an} & 18.7 & 86.5 & 384 & 77.9 & - & - \\
        & ViT-L/16~\cite{dosovitskiy2021an} & 65.8 & 304.33 & 384 & 76.5 & - & - \\
        \cmidrule{2-8}
        & DeiT-T~\cite{DBLP:journals/corr/abs-2012-12877} & 1.2 & 5.7 & 224 & 72.2 & 91.1 & 80.6 \\
        & DeiT-S~\cite{DBLP:journals/corr/abs-2012-12877} & 4.5 & 22.1 & 224 & 79.9 & 95.0 & 85.7 \\
        & DeiT-B~\cite{DBLP:journals/corr/abs-2012-12877} & 17.3 & 86.6 & 224 & 81.8 & 95.6 & 86.7 \\
        & DeiT-T + Teacher~\cite{DBLP:journals/corr/abs-2012-12877} & 1.2 & 5.7 & 224 & 74.5 & 91.9 & 82.1 \\
        & DeiT-S + Teacher~\cite{DBLP:journals/corr/abs-2012-12877} & 4.5 & 22.1 & 224 & 81.2 & 95.4 & 86.8 \\
        \cmidrule{2-8}
        & DeiT-B$\uparrow$384~\cite{DBLP:journals/corr/abs-2012-12877} & 52.8 & 86.6 & 384 & 83.1 & 96.2 & 87.7 \\
        \cmidrule{2-8}
        & T2T-ViT-14~\cite{DBLP:journals/corr/abs-2101-11986} & 5.2 & 21.5 & 224 & 81.5 & - & - \\
        & T2T-ViT-19~\cite{DBLP:journals/corr/abs-2101-11986} & 8.9 & 39.2 & 224 & 81.9 & - & - \\
        & T2T-ViT-24~\cite{DBLP:journals/corr/abs-2101-11986} & 14.1 & 64.1 & 224 & 82.3 & - & - \\
        \cmidrule{2-8}
        & PVT-T~\cite{DBLP:journals/corr/abs-2102-12122} & 1.9 & 13.2 & 224 & 75.1 & - & - \\
        & PVT-S~\cite{DBLP:journals/corr/abs-2102-12122} & 3.8 & 24.5 & 224 & 79.8 & - & - \\
        & PVT-M~\cite{DBLP:journals/corr/abs-2102-12122} & 6.7 & 44.2 & 224 & 81.2 & - & - \\
        & PVT-L~\cite{DBLP:journals/corr/abs-2102-12122} & 9.8 & 61.4 & 224 & 81.7 & - & - \\
        \cmidrule{2-8}
        & CeiT-T & 1.2 & 6.4 & 224 & \textbf{76.4} & 93.4 & 83.6 \\
        & CeiT-S & 4.5 & 24.2 & 224 & \textbf{82.0} & 95.9 & 87.3 \\
        \cmidrule{2-8}
        & CeiT-T$\uparrow$384 & 3.6 & 6.4 & 384 & \textbf{78.8} & 94.7 & 85.6 \\
        & CeiT-S$\uparrow$384 & 12.9 & 24.2 & 384 & \textbf{83.3} & 96.5 & 88.1 \\

        \bottomrule
    \end{tabular}
    \label{tab:sota}
\end{table*}

We report the results on ImageNet validation dataset and ImageNet Real dataset \cite{DBLP:journals/corr/abs-2006-07159} in Table \ref{tab:sota}. For comparison, we select CNNs (ResNets \cite{DBLP:conf/cvpr/HeZRS16}, EfficieNets \cite{DBLP:conf/icml/TanL19}, RegNets \cite{radosavovic2020designing}) and Transformers (ViTs, DeiTs) to evaluate the effectiveness of our CeiT models. 

\paragraph{CeiT vs CNNs.} We first compare CeiT models with CNN models. CeiT-T achieves a Top-1 accuracy of $76.4\%$ in ImageNet, which is close to the performance of ResNet-50. But CeiT-T only requires $3\times$ fewer FLOPs and $4\times$ fewer Params than ResNet-50. For the CeiT-S of a similar size as ResNet-50, its performance is $82.0\%$, achieving a higher performance ($+5.3\%$) than that of ResNet-50 ($76.7\%$). This performance also outperforms larger CNN models of ResNet-152 and RegNetY-8GF. When trained on the resolution of $384\times 384$, CeiT-S$\uparrow384$ surpasses EfficientNet-B4 by $0.4\%$. It shows that we have obtained comparable results with EfficientNets, and have almost closed the gap between vision Transformers and CNNs.

\paragraph{CeiT vs ViT/DeiT/T2T/PVT.} 
CeiT-T achieves a similar result of $76.4\%$ with ViT-L/16 of $76.5\%$. This is a surprising result since the size of the CeiT-T model is only one-fifth the size of ViT-L/16. But this result is produced by the improvements of the training strategy and the modifications of the model structure. To further demonstrate the improvements brought by the structure, we compare CeiT with DeiT. CeiT models follow the same training strategy as given in section \ref{sec:exp_setting}. Our modifications only increase the number of parameters by about $10\%$, and have almost no effect on FLOPs. In this way, CeiT-T outperforms DeiT-T by a large margin of $4.2\%$ for the Top-1 accuracy. And CeiT-S obtains higher results than that of DeiT-S and DeiT-B by $2.1\%$ and $0.2\%$ respectively. 
We also compare CeiT with concurrent work of T2T-ViT and PVT.
CeiT-S achieves slightly higher accuracy than T2T-ViT-19 with much fewer FLOPs and parameters.
For PVT models, CeiT-T obtains higher accuracy of $1.3\%$ than PVT-T with fewer FLOPs and parameters.
CeiT-S also outperforms PVT-S/M/L models separately.

\paragraph{CeiT vs DeiT-Teacher.} DeiT introduces a CNN teacher model as the extra supervision to optimize the Transformer, achieving higher performances. But it requires extra computation cost to obtain the trained CNN model. While CeiT does not need an additional CNN model to provide supervision information, except for the ground truth. Meanwhile, CeiT-T surpasses DeiT-T-Teacher by $1.9\%$ of the Top-1 accuracy. And CeiT-S also outperforms DeiT-S-Teacher by $0.8\%$. These experimental results demonstrate the effectiveness of our CeiT.

\subsection{Transfer Learning}\label{sec:exp_transfer}

To demonstrate the generalization power of pre-trained CeiT models, we conduct experiments of transfer learning in 7 downstream benchmarks. And the results are given in Table \ref{tab:transfer}. Training details are given in the previous Table \ref{tab:hyper}. It can be seen that CeiT-S outperforms DeiT-B in most datasets with fewer parameters and FLOPs. CeiT-S$\uparrow384$ achieves state-of-the-arts results in most datasets. Notably, CeiT-S$\uparrow384$ get comparable results with EfficientNet-B7 with an input size of 600. 
It suggests that the pretrained models of Transformers can be well migrated to downstream tasks, showing the strong potential of visual Transformers against CNNs.

\subsection{Ablation Studies}\label{sec:exp_ablation}

\begin{table}[t]
    \centering
    \footnotesize
    \caption{Ablation study results on the type of I2T. Top-1 accuracy and changes are reported.}
    \begin{tabular}{cccccc}
         \toprule
         \multicolumn{4}{c}{I2T Type} & \multirow{2}{*}{Top-1} \\
         \cmidrule{1-4}
         conv & maxpool & BN & channels & \\
         \midrule
         \xmark & \xmark & \xmark & 3 & 72.2 \\
         \midrule
         $k7s4$ & \xmark & \xmark & 64 & 71.4 (\textcolor{Green}{-0.8}) \\
         $k5s4$ & \xmark & \xmark & 64 & 71.1 (\textcolor{Green}{-1.1}) \\
         $k3s2$ + $k3s2$ & \xmark & \xmark & 64 & 70.4 (\textcolor{Green}{-1.8}) \\
         $k7s2$ & $k3s2$ & \xmark & 32 & 72.9 (\textcolor{Red}{+0.7}) \\
         $k7s2$ & $k3s2$ & \cmark & 32 & 73.4 (\textcolor{Red}{+1.2}) \\
         \bottomrule
    \end{tabular}
    \label{tab:i2t}
\end{table}
To further identify the effects of the proposed modules, we conduct ablation studies on the main components of I2T, LeFF, and LCA. All of our ablation experiments are based on the DeiT-T model on ImageNet.

\begin{table*}[t]
   \centering
   \footnotesize
   \caption{Results on downstream tasks with ImageNet pre-training. CeiT models achieve state-of-the-arts performance. The results with the first two highest accuracies are bolded.}
   \begin{tabular}{c|c|cccccccc}
       \toprule
       Model & FLOPs & ImageNet & iNat18 & iNat19 & Cars & Followers & Pets & CIFAR10 & CIFAR100 \\
       \midrule
       Grafit ResNet-50~\cite{touvron2020grafit} & 4.1G & 79.6 & 69.8 & 75.9 & 92.5 & 98.2 & - & - & - \\
       Grafit RegNetY-8GF~\cite{touvron2020grafit}  & 8.0G & - & 76.8 & 80.0 & 94.0 & \textbf{99.0} & - & - & - \\
       EfficientNet-B5~\cite{DBLP:conf/icml/TanL19} & 10.3G & 83.6 & - & - & - & 98.5 & - & 98.1 & \textbf{91.1} \\
       EfficientNet-B7~\cite{DBLP:conf/icml/TanL19} & 37.3G & \textbf{84.3} & - & - & \textbf{94.7} & \textbf{98.8} & - & 98.9 & \textbf{91.7} \\
       \midrule
       ViT-B/16~\cite{dosovitskiy2021an} & 18.7G & 77.9 & - & - & - & 89.5 & 93.8 & 98.1 & 87.1 \\
       ViT-L/16~\cite{dosovitskiy2021an} & 65.8G & 76.5 & - & - & - & 89.7 & 93.6 & 97.9 & 86.4 \\
       \midrule
       Deit-B~\cite{DBLP:journals/corr/abs-2012-12877} & 17.3G & 81.8 & 73.2 & 77.7 & 92.1 & 98.4 & - & \textbf{99.1} & 90.8 \\
       Deit-B$\uparrow$384~\cite{DBLP:journals/corr/abs-2012-12877} & 52.8G & 83.1 & \textbf{79.5} & \textbf{81.4} & 93.3 & 98.5 & - & \textbf{99.1} & 90.8 \\
       \midrule
        CeiT-T & 1.2G & 76.4 & 64.3 & 72.8 & 90.5 & 96.9 & 93.8 & 98.5 & 88.4 \\
        CeiT-T$\uparrow$384 & 3.6G & 78.8 & 72.2 & 77.9 & 93.0& 97.8 & 94.5 & 98.5 & 88.0\\ 
        CeiT-S & 4.5G & 82.0 & 73.3 & 78.9 & 93.2 & 98.2 & \textbf{94.6} & 99.0 & 90.8 \\
        CeiT-S$\uparrow$384 & 12.9G & \textbf{83.3} & \textbf{79.4} & \textbf{82.7} & \textbf{94.1} & 98.6 & \textbf{94.9} & \textbf{99.1} & 90.8 \\
       \bottomrule
   \end{tabular}
   \label{tab:transfer}
\end{table*}

\paragraph{Different types of I2T module.} The influencing factors in I2T include the kernel size of the convolution, the stride of the convolution, the existence of Max-pooling and BatchNorm layers. The results are given in Table \ref{tab:i2t}.  Without the Max-pooling layer, one convolution layer with a kernel of $k7s4$ and $k5s4$ each decreases the performance. An I2T with two convolution layers with a kernel of $k3s2$ also suffers from a drop. Both the Max-pooling and BatchNorm layers benefit the training. Therefore, we adopt the best structure (in the last row) in all of our experiments.

\begin{table}[t]
    \centering
    \footnotesize
    \caption{Ablation study results on the type of LeFF. Top-1 accuracy and changes are reported.}
    \begin{tabular}{ccc}
         \toprule
         \multicolumn{2}{c}{LeFF Type} & \multirow{2}{*}{Top-1} \\
         \cmidrule{1-2}
         kernel size & BN & \\
         \midrule
         \xmark & \xmark & 72.2 \\
         \midrule
         $1\times 1$ & \xmark & 70.3 (\textcolor{Green}{-1.9}) \\
         $3\times 3$ & \xmark & 72.7 (\textcolor{Red}{+0.5}) \\
         $5\times 5$ & \xmark & 73.1 (\textcolor{Red}{+0.9}) \\
         \midrule
         $3\times 3$ & \cmark & 74.3 (\textcolor{Red}{+2.1}) \\
         $5\times 5$ & \cmark & 74.4 (\textcolor{Red}{+2.2}) \\
         \bottomrule
    \end{tabular}
    \label{tab:leff}
    \vspace{-1.5em}
\end{table}

\paragraph{Different types of LeFF module.} In a LeFF module, the size of the kernel determines the region size in which patch tokens establish local correlation. So we test using kernel sizes of $1\times 1$, $3\times 3$ and $5\times 5$ in Table \ref{tab:leff}. Compared to the baseline without the middle depth-wise convolution, the type of $1\times 1$ shows poor performance with a drop of $1.9\%$. This shows that simply increasing the number of layers for the Transformer does not certainly bring improvements. When increasing the kernel size to larger ones, each token can accumulate with neighboring tokens through the non-linear transformation. Both the types of $3\times 3$ and $5\times 5$ obtain gains. When adopting the BatchNorm layer, the model can achieve further accuracy improvements up to $2.2\%$ of Top-1 accuracy. Based on the trade-off between the number of parameters and accuracy, we choose the kernel size of $3\times 3$. The same as I2T, the presence of BatchNorm layers following transformation layers significantly improves the performance.

\paragraph{Effectiveness of LCA.} We compare the performances w/wo the LCA module. Through adopting LCA, the performance improves from $72.2\%$ to $72.8\%$, showing multi-level information contributes to the final image representation.

\begin{table}[t]
   \centering
   \footnotesize
   \caption{Comparisons of the ability of convergence between DeiT and CeiT models. CeiT models trained with 100 epochs obtain comparable results with DeiT models trained with 300 epochs. $1\times$ means 100 epochs.}
   \begin{tabular}{cc|cc|cc}
       \toprule
       $3\times$ & Top-1 & $1\times$ & Top-1 & $1\times$ & Top-1\\
       \midrule
       DeiT-T & 72.2 & DeiT-T & 65.3 & CeiT-T & 72.2 (\textcolor{Red}{+6.9}) \\
       DeiT-S & 79.9 & DeiT-S & 74.5 & CeiT-S & 78.9 (\textcolor{Red}{+4.4}) \\
       DeiT-B & 81.8 & DeiT-B & 76.8 & CeiT-B & 81.8 (\textcolor{Red}{+5.0}) \\
       \bottomrule
   \end{tabular}
   \label{tab:convergence}
\end{table}

\subsection{Fast Convergence}\label{sec:exp_fast}

The standard visual Transformers, such as ViT and DeiT, usually require a large number of training epochs to converge. Using $3\times$ fewer training epochs, the performances of DeiT suffer significant declines. As shown in Table \ref{tab:convergence}, CeiT models demonstrate better convergence than DeiT models, resulting in higher performances in a large margin. And CeiT models trained in 100 epochs can obtain comparable results with DeiT models trained in 300 epochs. It shows that incorporating these inductive biases inherent in CNNs benefits the optimization procedure of visual Transformers. 

\section{Conclusion}

In this paper, we propose a new CeiT architecture that combines the advantages of CNNs in extracting low-level features, strengthening locality, and the advantages of Transformers in establishing long-range dependencies. CeiT obtains state-of-the-art performances on ImageNet and various downstream tasks, without requiring a large amount of training data and extra CNN teachers. Besides, CeiT models demonstrate better convergence than pure Transformer with $3\times$ fewer training iterations, reducing the training cost significantly. Through incorporating convolution designs, we provide a new perspective for more effective visual Transformers.

{\small
\bibliographystyle{ieee_fullname}
\bibliography{egbib}
}

\end{document}